%% file: main.tex
\documentclass[10pt, a4paper]{article}

\usepackage{booktabs}
\usepackage{multirow,graphicx}
\usepackage[final]{lrec2026} 

\title{Evaluating the Homogeneity of Keyphrase Prediction Models}

\name{Maël Houbre$^1$, Florian Boudin$^2$, Béatrice Daille$^3$} 

\address{$^1$Ministerial Agency of Artificial Intelligence in Defense, France \\
        $^2$Inria, LS2N, Nantes Université, France \\
        $^3$LS2N, Nantes Université, France \\
        mael.houbre@def.gouv.fr, \{florian.boudin, beatrice.daille\}@univ-nantes.fr}

\abstract{
Keyphrases which are useful in several NLP and IR applications are either extracted from text or predicted by generative models.
Contrarily to keyphrase extraction approaches, keyphrase generation models can predict keyphrases that do not appear in a document's text called `absent keyphrases`. This ability means that keyphrase generation models can associate a document to a notion that is not explicitly mentioned in its text. 
Intuitively, this suggests that for two documents treating the same subjects, a keyphrase generation model is more likely to be homogeneous in their indexing i.e. predict the same keyphrase for both documents, regardless of those keyphrases appearing in their respective text or not; something a keyphrase extraction model would fail to do. 
Yet, homogeneity of keyphrase prediction models is not covered by current benchmarks. In this work, we introduce a method to evaluate the homogeneity of keyphrase prediction models and study if absent keyphrase generation capabilities actually help the model to be more homogeneous. 
To our surprise, we show that keyphrase extraction methods are competitive with generative models, and that the ability to generate absent keyphrases can actually have a negative impact on homogeneity. Our data, code and prompts are available on hugginface and github.
\\ \newline  }

\begin{document}

\maketitleabstract

\section{Introduction}

Keyphrases are expressions that represent the key aspects and notions of a document. They are useful in a variety of NLP tasks, such as summarization or classification, but also for information extraction, such as document indexing.

Predicting keyphrases can be done either through extraction or generation. In that last case, keyphrase generation models are able to generate keyphrases that do not appear in the document (called `absent keyphrases`). This ability to generate absent keyphrases is one of the reasons why generation models outperform extraction model in intrinsic and extrinsic evaluations, especially in the latter, as generating absent keyphrases indeed brings additional information and thus improves the document's indexing\citep{boudin_redefining_2021}.
However, the evaluation of keyphrase prediction models is still an under-studied field of research as the vast majority of approaches are still only evaluated by comparing with reference keyphrases. This kind of evaluation only measures the model's ability to reproduce the annotation of reference, not the quality of the predicted keyphrases. 

When it comes to document indexing, \citet{firoozeh_keyword_2020} detailed several properties that keyphrases or keyphrase prediction models must satisfy to be useful. However, only a minority of these properties are commonly evaluated. One of those properties that has yet to be studied is homogeneity, i.e. the ability to consistently assign the same keyphrase to the same concept in different documents. This property is extremely important to ensure the consistency of the indexing process. 
\citet{rolling_indexing_1981} presented the lack of homogeneity in the indexing as one of the limitations of having multiple human indexers. The more indexers, the more difficult it becomes to have the same keyphrase attributed to the same notion in different documents. Automatically indexing  with a keyphrase prediction model might be a solution to that problem, since all the indexing is done by a single "annotator". However, the homogeneity of the indexing from that kind of models is not guaranteed. Indeed, contrarily to controlled indexing, which performs indexing with a thesaurus and thus seeks to guaranty homogeneity, keyphrase prediction models are almost exclusively trained in a free indexing setup (i.e. without vocabulary constraints).

In this work, we propose a method to evaluate the homogeneity of keyphrase prediction models in free indexing. To that end, we construct pairs of documents that have the same key aspects and measure the similarity of the predicted keyphrases for both documents . Our pairs are from documents from widely used datasets in the computer science domain.

Our contributions are as follows. We introduce a method to evaluate the homogeneity of keyphrase prediction models, thus widening the panel of studied quality properties. We thoroughly analyse our results and show that, to our surprise, absent keyphrase generation abilities can actually have a negative impact on homogeneity in a free indexing setup. We make our data pairs and our code available to the community on huggingface and github respectively.

\section{Hypothesis and approach}

\citet{shen_unsupervised_2022} discovered that in the Inspec collection~\citep{hulth_improved_2003}, a widely used dataset of bibliographic records in the computer science domain annotated by professional indexers, 95\% of the absent keyphrases were present keyphrases in other documents of the same collection. This means that for two documents having common concepts, there is a chance that the associated keyphrases will not be present in the source text of both documents.

As only keyphrase generation models have the ability to generate absent keyphrases, keyphrase extraction approaches would then fail to be consistent in indexing both documents. Therefore, our hypothesis is that keyphrase generation models can produce more homogeneous predictions than keyphrase extraction models, and that mechanisms improving generation capabilities must also improve homogeneity.

\section{Experimental Settings}

\subsection{Evaluation metrics}

\begin{figure*}
    \centering
    \includegraphics[width=\linewidth]{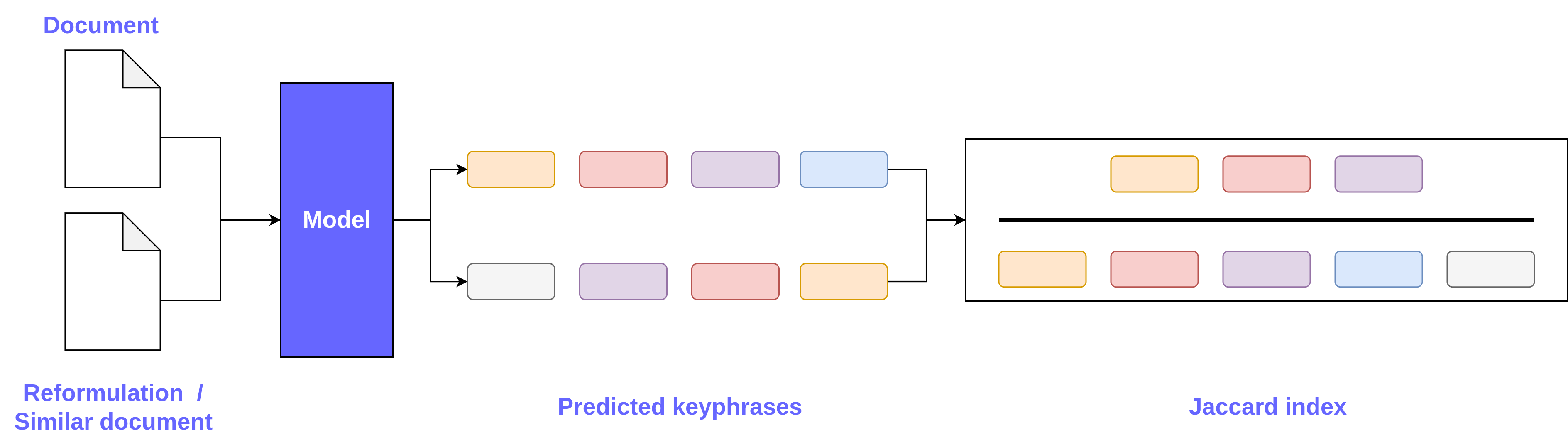}
    \caption{Illustration of our evaluation approach. Keyphrases, represented as coloured rectangles, are predicted for both articles and the intersection between the predictions is measured.}
    \label{fig:protocol_fig}
\end{figure*}

When we consider the keyphrase prediction model as an annotator for an indexing task, the definition of homogeneity is closely related to the definition of intra- and inter-annotator consistency~\citep{zunde_indexing_1969, leonard_inter-indexer_1977,rolling_indexing_1981} i.e. we evaluate whether the model is consistent in its indexing of same concepts from different documents. Thus, we can apply a measure of intra-annotator consistency on pairs of similar documents to evaluate homogeneity. We consider that for two documents of the same topics, a homogeneous keyphrase prediction model must predict the same keyphrases or at least, similar ones. Our evaluation protocol is illustrated in figure~\ref{fig:protocol_fig}. For each pair of documents, we predict keyphrases on both documents and then measure the consistency of both predictions.

To measure the homogeneity, we use two metrics from intra- and inter-annotator consistency studies: Hooper's CP and Rodgers' CP where "CP" stands for "consistency of a pair"~\citep{leonard_inter-indexer_1977}. These two metrics are Jaccard index (i.e. intersection over union) applied on keyphrase sequences. Hooper's CP measures the intersection over union at the keyphrase scale while Rodgers' CP, a more permissive metric, measures it on a word scale. Let's say a model predicts "\textit{natural language processing, homogeneity evaluation, keyphrase generation}" for the first document and "\textit{natural language processing, homogeneity evaluation, text generation}" for the second document. For those two sequences, the intersection at the keyphrase level (i.e. Hooper's CP) will be 2 keyphrases out of 3, and at the word level (i.e. Rodgers' CP), it will be 6 unique words out of 7.



\subsection{Models}

We conduct our evaluation on 8 keyphrase generation models as well as 2 keyphrase extraction approaches as baselines. We chose those 8 generative models as we think that they represent a good overview of the different works on keyphrase generation architectures since the introduction of keyphrase prediction as a sequence generation task.

\begin{itemize}
    \item \textbf{CopyRNN}~\citep{meng_deep_2017}: the first encoder-decoder model applied for keyphrase generation. The model employs a copy mechanism, allowing it to copy words from the source text to tackle the problem of keyphrases composed of out of vocabulary words. We also evaluate a variant of this model, CorrRNN~\citep{chen_keyphrase_2018} which adds a coverage attention mechanism to improve the diversity of the predictions.
    
    \item \textbf{TG-Net}~\citep{chen_title-guided_2019}: an encoder-decoder model with a copy mechanism but with an additional encoding, matching parts of the title to the rest of the text in order to give it more weight in the overall text representation. We also use another version of this model, TGRF~\citep{chan_neural_2019} is the same architecture but trained with a reinforcement learning policy.
    
    \item \textbf{BART}~\citep{lewis_bart_2020}: a pre-trained encoder-decoder model that has been widely used for keyphrase generation~\citep{chowdhury_applying_2022, kulkarni_learning_2022, meng_general--specific_2023, boudin_unsupervised_2024, houbre_self-compositional_2025} as it has shown state of the art performances on summarization tasks to which keyphrase generation is sometimes considered a related task.

    \item \textbf{One2Set}~\citep{ye_one2set_2021}: the current state of the art architecture for keyphrase generation which generates keyphrases in a set of independent keyphrases rather than a sequence~\citep{ye_one2set_2021, xie_wr-one2set_2022}

    \item \textbf{MultipartiteRank}~\citep{boudin_unsupervised_2018}: an unsupervised keyphrase extraction method relying on a multipartite graph and topic modelling to improve the coverage of a document's topics.

    \item \textbf{TF-IDF}~\citep{salton_vector_1975}: an unsupervised method that ranks candidate keyphrases based on the ratio of their frequency in the document and their frequency in the overall corpus.

\end{itemize}

For BART and One2Set models, we also train those models, using the self-compositional data augmentation method from~\citet{houbre_self-compositional_2025} which increases the original set of documents with synthetic generated documents. As this method has been shown to improve absent keyphrase generation, we want to evaluate if indexing a document with a method that favours the generation of absent keyphrases has a positive impact on homogeneity. We refer to those models as BART-60p and One2Set-60p respectively.

All generative models are trained on KP20k\citep{meng_deep_2017} which is the most widely used dataset for keyphrase generation. It comprises 530k bibliographic records annotated by the authors. At the exception of One2Set which has its own training paradigm, all generative models are trained on the One2Seq paradigm (i.e. all the keyphrases in a single sequence) and in an auto terminating setting. This means that the model has to determine how many keyphrases it has to predict for a given document~\citep{yuan_one_2020}. The keyphrase sequences for training are ordered first with present keyphrases in their apparition order in the source text, followed by absent keyphrases in their original order. We chose this order as it had the best performances according to~\citet{meng_empirical_2021}. For the CopyRNN model, CorrRNN, TG-Net and TGRF, we train them with the same parameters as from~\citet{chan_neural_2019} and use the code made available by the authors~\footnote{\url{https://github.com/kenchan0226/keyphrase-generation-rl}}.

For the keyphrase extraction models, we use the implementation available in the pke library~\citep{boudin:2016:COLINGDEMO}. For those methods, the tool only allows us to predict a fixed number of keyphrases for each document. As keyphrase generation models have been trained on KP20k, a dataset with an average of 5 keyphrases per document, we fix the number of predictions for MultipartiteRank and TF-IDF to 5.

\subsection{Getting evaluation data}

\input{Tables/eval_data_stats}

To evaluate homogeneity, we need pairs of documents that treat the same notions. We identify two ways to achieve this goal. The first one is to reformulate a document to have a second version. The second one is making pairs with documents that share keyphrases, under the hypothesis that documents sharing keyphrases are semantically similar. The statistics of our datasets are available in Table~\ref{ch6_tab:eval_data_stats} with the proportion of keyphrases according to the PRMU classification paradigm~\citep{boudin_redefining_2021}.

\subsubsection{Pairs with reformulation}

Getting a second version of a document through reformulation has the advantage of preserving the overall meaning and therefore, the pertinence of the assigned keyphrases. There are several ways to reformulate a document. We can replace some words by synonyms, generate paraphrase or use back-translation. 
As large language models have shown good performances in paraphrase and back-translation~\citep{witteveen_paraphrasing_2019, wang_benchmarking_2024}, we opt for those two methods. We select two methods to make sure that we have enough evaluation examples, even after filtering reformulations of lesser quality. 

For those two tasks, we use the GPT-4o model~\citep{openai_gpt-4_2024} as it is one of the best performing model for text generation tasks at the moment of this work. As our data is in English, we choose the English-Spanish language pair for back-translation. 
The choice of this language pair comes from the hypothesis that as GPT models have been trained on web data, it is very likely that they have been trained on popular massive datasets such as Oscar~\citep{ortiz_suarez_monolingual_2020}. On this dataset, the two most represented languages are English and Spanish. Also, when asking GPT-4o to give the two languages it has been trained on the most, the answer was English and Spanish.

Once the reformulations are  generated, we ensure and evaluate their quality upon two criteria, monotony and diversity. To ensure the quality of the reformulation, we check that the present keyphrases of the source document all occur in the reformulation. Indeed, as keyphrases represent the key aspects of a document, notions that they carry should not be lost in the reformulation. Occurrences of present keyphrases of the source document are either full matches (i.e. they are still present keyphrases) or have all their constitutive words in the text but not in a contiguous sequence(i.e. they are `reordered' keyphrases according to the PRMU keyphrase classification paradigm~\citep{boudin_redefining_2021})
Another criteria that could affect reformulation quality is the variation in word count. Having a reformulation that is too short may be a sign of information loss whereas a longer one may be synonym of noise introduction. Therefore we only keep reformulations for which the word count is within 10~\% of the original text's.

To create our reformulation pairs, we employ the Inspec dataset~\citep{hulth_improved_2003}. This dataset includes 500 bibliographic records with keyphrases assigned by professional indexers. As documents in this dataset are assigned a high number of keyphrases (around 10 per document) and that most of them (79\% on average) are present keyphrases, we think that this will result with only high quality reformulations after filtering. 

Back-translation and paraphrasing give us 1\,000 reformulated documents. After applying our filters, we obtain 280 reformulations that meet our quality criteria (170 from back-translation and 110 from paraphrasing). Those synthetic documents have 68.2\% of Hooper's CP score which means that a perfectly homogeneous keyphrase extraction model can only reach 68.2\% of Hooper's CP score, while a perfectly homogeneous keyphrase generation model should get the Hooper score of 100\%.

Our last checking of reformulation quality is about diversity.
As the synthetic documents are obtained through reformulation, we check that they sufficiently varied in forms.
For that we use two metrics, the word error rate (WER) and ROUGE-1. 
The word error rate is a metric mainly employed in speech to text and is an adaptation of the Levenshtein distance which measures the amount actions necessary to go from the prediction (here our reformulation) back to the reference (i.e. the original document). This metric does not measure the amount of information in common between both texts, but rather the amount of morphological modifications (word order, etc.) that has been applied to the original document. 
However, even if the original text must be modified, we have to make sure that the information contained in the original text has been preserved. Even if our filter on the preservation of present keyphrases already focuses on that, all the information may not be contained in present keyphrases. Therefore we use ROUGE-1 which will compute the recall of the words from the original document in the reformulation. We can say that this metric measures how similar the vocabularies of both documents are. A good reformulation should then have both high WER and high ROUGE-1 scores. 

On average, the reformulations have a WER of 45~\% and a ROUGE-1 score of 66~\%. The WER score means that on average, a bit less than half the words from the original documents have either been deleted, moved or substituted for another word. Yet, the ROUGE-1 score shows that on average, two thirds of the vocabulary of the original document is preserved. Those scores paired with our quality filters show that even if there has been a significant amount of modifications, most of the information from the original document has been preserved.

\input{Tables/inspec_results}

\subsubsection{Pairs from shared keyphrases}

One limitation of evaluating with reformulations is that the quality of the evaluation process is utterly reliant on the quality of the reformulation. One alternative is to obtain evaluation pairs selected from the set of source texts. To solve this issue we leverage shared keyphrases among documents to measure their similarity~\citep{houbre_self-compositional_2025}. As keyphrases illustrate the main concepts of a document, we consider that the more keyphrases two documents share, the more semantically similar they are.

\begin{equation}
    sim = \frac{|A \cap B|}{max(card(A),card(B))}
    \label{equation_houbre}
\end{equation}

\citet{houbre_self-compositional_2025} measured the similarity of documents as represented in Equation~\ref{equation_houbre} where A and B are keyphrase sequences. In their work, the minimum similarity value was fixed to 60~\%. As the Jaccard index is already close to that definition (i.e. with the intersection of A and B rather than the maximum of cardinal), we employ the Jaccard index at the keyphrase level to measure the similarity between document pairs.

The low amount of documents from the Inspec test set does not allow us to apply this method on it. Indeed, even with a similarity value as low as 20\%, we only get 16 document pairs. To tackle this issue, we build our document pairs on the test set of KP20k which has 20~000 documents. With this dataset and a constraint of a minimum similarity value of 50\%, we obtain 145 evaluation pairs from 225 unique documents. This means that an homogeneous keyphrase prediction model should display a Hooper's CP of at least 50\%.

We see in Table~\ref{ch6_tab:eval_data_stats} that contrarily to the evaluation pairs made from reformulations, the pairs obtained from shared keyphrases have a slight majority of absent keyphrases; from those absent keyphrases, unseen keyphrases i.e. with none of their constitutive words in the source text, are the most represented. These statistics suggest that on this dataset, keyphrase generation models with good performances on absent keyphrases should be challenging. 

\section{Results and analysis}

The results for both scenarios are detailed in Tables~\ref{tab:inspec_results} and~\ref{tab:kp20k_results}. For Table~\ref{tab:inspec_results}, we also add the Spearman correlation coefficient between the Hooper's CP and the ROUGE-1 score between a document and its reformulation. The goal is to verify if keyphrase generation models are less sensitive to variation between texts than keyphrase extraction models.

\subsection{Results on the pairs from reformulations}

One of the first result is that keyphrase prediction models are not homogeneous. Indeed, TGRF, which is the best performing model, only has a Hooper's score of 57.6~\% on the pairs from the Inspec evaluation set. This means that even for the best performing model, only slightly more than half of the predicted keyphrases for a document pair are the same.

To check if the models that generate the most absent keyphrases are the most homogeneous, we add the proportion of absent keyphrases generated for each model. We remark that the best performing generative model (TGRF) is actually the one that generates the least amount of absent keyphrases. This suggests that contrarily to our hypothesis, generative capabilities do not necessarily mean better homogeneity. This suggestion is strengthened by two other observations. Firstly, the model with the worst homogeneity score of 29.6 (One2Set) is the model the second highest proportion of absent keyphrases (40.2\%), close behind the score of its version with data augmentation One2Set-60p (41.1\%) . Secondly, we see that TF-IDF, the best performing keyphrase extraction method, ranks second on homogeneity scores, outperforming 7 of the 8 keyphrase generation methods. MultipartiteRank also outperforms the majority of keyphrase generation methods (5 out of 8). Those results contradict our main hypothesis as generative models are shown to be less homogeneous than extractive methods.

This tendency trend appears also among models with the same architecture. TG-Net and TGRF share  the same architecture but with a different training loss. Yet, TG-Net which generates more absent keyphrases than TGRF (28\% against 11.7\%) on our examples, has much lower homogeneity scores with a Hooper's CP of 45.7\% when TGRF has 57.6\%.
BART architecture shows similar behaviour. Even if the base model and the one trained with the self-compositional data augmentation method have similar absent keyphrase generation rates (26.6~\% against 26.1~\%), the latter has slightly lower similarity scores (37.3 against 37.9 in Hooper's CP) . But as mentioned by \citet{houbre_self-compositional_2025}, the data augmentation method resulted in improved performance on absent keyphrases. These results suggest that the generation of absent keyphrases has a negative impact on homogeneity.

To verify if the amount of modifications in the reformulations are correlated with the homogeneity (or the lack thereof), we add to Table~\ref{tab:inspec_results} the Pearson correlation factor between the homogeneity score and the ROUGE-1 between the source document and its reformulation.
The values of this column show that for most models, there is a slight positive correlation between the ROUGE-1 of an original document and its reformulation and the homogeneity score of the model. Without surprises, extractive models are the most sensitive to lexical variations between two versions of a text. However, for almost all the generative models, the correlation coefficient is weak (inferior to 0.3). Meaning that we cannot consider the variations from the reformulation as the only cause for the low homogeneity scores.

One hypothesis for the lower homogeneity scores for generative model is that their wide vocabulary may deteriorate homogeneity. Indeed, with extractive methods, the vocabulary is contained within the source text which for bibliographic records, is only several hundreds of words long. With generative models, they generate keyphrases by attributing a probability distribution on their vocabulary which is most of the time composed of several tens of thousands of words or tokens \citep{meng_deep_2017,lewis_bart_2020,ye_one2set_2021}. Generating keyphrase with a wide vocabulary in a free indexing setting may therefore introduce variability in the generation which then has a negative impact on homogeneity.

The only exception to that tendency seems to be the One2Set-60p architecture. As showed in Table~\ref{tab:inspec_results}, even if the One2Set-60p model  has the highest absent keyphrase generation rate (41.1\%), it ranks fourth on Hooper's CP with 48.2\%. Also, even if its generation rate is analogue to One2Set (41.1\% against 40.2\%), One2Set is actually the worst performing model with 29.6\% of Hooper's CP. Our hypothesis for this difference in behaviour compared to other generation models lies in One2Set's training paradigm. When training the model to generate a set of independent keyphrases, \citet{ye_one2set_2021} operates a bipartite pairing between the predictions and the reference. The model then learns the link between predictions and reference twice: first with the bipartite pairing and another time with the loss function. As the synthetic documents from the augmentation method are obtained from combinations of documents from the original training set, they may have improved the bipartite pairing, hence improving homogeneity. Another possibility is that the data augmentation method  have strengthened the relation between a context and a keyphrase, making the augmented model more homogeneous. This could explain why the augmented model has the highest correlation of 0.37 between ROUGE-1 and homogeneity score amongst generative approaches.

\input{Tables/kp20k_results}

\subsection{Results on pairs with shared keyphrases}

The results on the pairs from shared keyphrases are displayed in Table~\ref{tab:kp20k_results}. The first result is that the keyphrases extraction models, TF-IDF and MultipartiteRank, are the worst performing approaches.

The second result is that on the contrary of the results in the precedent section, models with more generative behaviours are shown to have better homogeneity scores. Indeed, the Hooper's score for BART-60p is 34.6\% better (+3.6 points) than the base BART model. The tendency is also observed for the One2Set architecture (+17\%) and the Title guided models TGRF and TG-Net (+17.9\% for the latter). One this test set, these results corroborate our hypothesis that generative models aught to be more homogeneous than keyphrase extraction models.

\subsection{Different reference sets, different results}
The two test sets that this work provides have different features which allows to evaluate different aspects of homogeneity for keyphrase prediction models. On one hand the reformulation pairs have a clear majority of present keyphrases and, through our filtering heuristics, have a significant lexical overlap (66\% Of ROUGE-1 on average). This set evaluates the homogeneity of keyphrase prediction models on lexically similar texts, which may give an edge to extraction models. On the other hand, the pairs from shared keyphrases have a slight majority of absent keyphrases, with most of them being keyphrases with some or all of their constitutive keyphrases not in the source text (i.e. `mixed` and `unseen` keyphrases according to the PRMU classification paradigm). This test set evaluates the homogeneity of keyphrase prediction models on documents sharing notions often not appearing in the text (i.e. abstractive concepts), giving an advantage to keyphrase generation models.

As the documents from KP20k are annotated by the authors, we also think that the document pairs with shared keyphrases have more generic keyphrases than the document pairs obtained from reformulations. Indeed, as authors tend to focus on potential user queries to choose their keyphrases~\citep{neveol_author_2010} they are likely to sometimes refer to broad topics when professional indexers focus on specific and discriminative notions. The link between both documents would therefore be more implicit than in the reformulation collection where documents share most of their key notions through present and reordered keyphrases. When manually looking at a small amount of random examples, we indeed observe that many shared keyphrases are very broad such as "design", "performance" or "algorithms".

To verify if this is a global characteristic, we compute the document frequency of each shared keyphrases among our document pairs as well as their length. We first consider that if a keyphrase is generic, it may be assigned to a large number of documents.  Then, we make the hypothesis that the longer a keyphrase is, the more specific it is. If we take this hypothesis the other way around, then a shorter keyphrase would be more generic. The keyphrases in the original test set of KP20k are on average assigned to 2 documents. When it comes to the shared keyphrases in our document pairs, they are assigned to an average of 24 documents, making them very common keyphrases compared to the rest of the KP20k collection. The same goes with the length of the keyphrases. On average, the keyphrases of the test set of KP20k are 1.9 words long, but only 1.7 words for the shared keyphrases in our document pairs. Those results show that the pairs are made from generic keyphrases. 

As generic keyphrases do not bring discriminative information about a document, the link between two documents from those pairs can be considered more implicit and therefore, more difficult to generate. This may explain why models such as BART-60p and One2Set-60p that have been trained with artificial data created with a similar setting are more homogeneous than their base model. 

The results that we have obtained on both test collections suggest that for two documents treating the same topics, choosing the best model to obtain the same keyphrases depends more on how explicit the link between those two documents is. When the link is explicit (i.e. there is a significant lexical overlap between the two documents), a keyphrase extraction method is at an advantage. However when the link is more implicit, that the keyphrases have a broader meaning, a model able to generate absent keyphrases seems more appropriate. As it has already been highlighted by \citet{boudin_keyphrase_2020}, this also suggests that present and absent keyphrases are somewhat complementary and can both bring information that can be beneficial to indexing.
Therefore, extracting keyphrases first and using them as an additional context to guide the generation of absent keyphrases intuitively seems to be a promising approach to address the lack of homogeneity of the approaches studied in this work.
These "hybrid" approaches have been explored in some previous works~\citep{chen_integrated_2019,ahmad_select_2021,wu_unikeyphrase_2021}. However, they represent a minority compared to approaches exploiting generative models for the full prediction.

\section{Related Work}

This work aims to broaden the coverage of keyphrase quality properties in automatic evaluation.

\subsection{Qualitative analysis of keyphrases}

As keyphrases represent the key aspects of a document,~\citet{firoozeh_keyword_2020} focused their work on defining the notion of "keyness" and discussed how keyphrase extraction models targeted certain features to represent that "keyness". The article presented 10 different properties regrouped in 3 categories: informational, linguistic and domain based. Linguistic properties verify that the keyphrases are linguistically sound and are in a form without inflexions. Informational properties are focused on whether the keyphrases are relevant and specific enough to be discriminative and cover all the aspects of a document without redundancies. Finally, domain-based properties are to determine if the keyphrases are selected from the domain's terminology and are unambiguous. The homogeneity property is also defined in this last category as it measures whether a keyphrase extraction method is consistent in choosing keyphrases for documents from the same domain. The article also underlines that many of those properties are still not taken into account in current evaluations.

\subsection{Evaluation of keyphrase prediction}

We distinguish two types of evaluations for keyphrase prediction: intrinsic and extrinsic. Intrinsic evaluation refers to evaluation considering keyphrase prediction as an end in itself. When the evaluation focuses on the impact of the predicted keyphrases for other tasks, we call it an extrinsic evaluation.

\paragraph{Intrinsic evaluation}

Intrinsic evaluation of keyphrase prediction is mainly focused on perfect match between the prediction and the reference, usually after stemming ~\citep{turney_learning_2000,hulth_improved_2003,meng_deep_2017,thomas_improving_2024}. The F-measure is the most employed metric, either on a fixed number of predictions such as 5 or 10 keyphrases~\citep{} or more recently, with variants adapting to predicted keyphrases sequences of different lengths~\citep{yuan_one_2020}. Even with stemming, one drawback of this kind of evaluation is that is considers as a wrong prediction, any keyphrase that does not exactly match the reference keyphrases. For example, predicting "neural networks" when the reference keyphrase is "deep neural networks" would be considered wrong. Some works have therefore included a more permissive evaluation relying on partial matching at different levels such as tokens, ngrams and lemmas~\citep{zesch_approximate_2009}. However as those evaluations are still at the lexical level, they may fail to account for semantically equivalent keyphrases that only have slight lexical overlap or even none at all.

With the use of word embeddings and more generally encoding models, comparing keyphrases using semantic distances allows us to measure semantic similarity between keyphrases. KPEval~\citep{wu_kpeval_2024} evaluates four different aspects of predicted keyphrases with this approach: the semantic similarity with the reference, the lack of redundancies of the predicted keyphrases, their faithfulness and their "utility" in a extrinsic setup.

The semantic similarity measure first uses a sentence-transformer model to encode the keyphrases. Then the cosine similarity is computed between each prediction and each reference and only the best matches for each are kept. The paper then introduces SemP, SemR and SemF, three measures relying on the definition of precision, recall and f-measure with the aforementioned best similarity matches. To evaluate the absence of redundancies, the same metrics are used but computed on each prediction relatively to the other predictions from the set. This measures how semantically similar the predictions are.

For the evaluation of faithfulness, KPEval employs a question answering model where the model is given a predicted keyphrase and a bibliographic record and is asked whether the prediction is consistent with the document's content. 

\paragraph{extrinsic evaluation}

The main work on extrinsic evaluation for keyphrase prediction is~\citet{boudin_keyphrase_2020}. They re-indexed documents with their content augmented with predicted keyphrases and evaluated a document retrieval model on the collection with augmented text and the one without. The authors then measured whether adding keyphrases had a beneficial impact or a negative impact on document retrieval. Another version of this work was presented in~\citet{boudin_redefining_2021} with a more detailed paradigm on absent keyphrases: the PRMU classification paradigm. Those works showed that adding author assigned keyphrases and keyphrases predicted by a model could improve document retrieval performances. When it comes to the distinction between present and absent keyphrases, the articles showed that even if present and absent keyphrases had both something to bring to document retrieval, absent keyphrases and more specifically keyphrases with parts completely absent from the source text, had a more significant impact. Both works were based on an external scientific collection NTCIR-2~\citep{kando_overview_2001}. KPEval adapted this approach by applying it on KP20k and KPTimes test sets. Requests corresponding to only one document each were created using a large language model and rather than re-indexing the whole documents, pairs of titles and predicted keyphrases were used.

\section{Conclusion}

In this work, we have introduced a method to automatically evaluate a new property of keyphrase prediction models: the homogeneity, i.e. the ability to consistently associate the same keyphrase to the same topic for different documents. This property is crucial to assess the quality of automatic document indexing and had yet to be evaluated.

We created two collections of similar documents in the computer science domain, one using reformulations of the original documents with a large language model and one with associations through shared keyphrases. Our experiments showed that when it comes to lexically similar documents, keyphrase extraction methods are superior to most of the keyphrase generation models. However, when the relation between documents is semantic, on a more generic and abstract level, the ability of generative models to generate absent keyphrases gives them an advantage over extractive approaches.

We hope that this work will encourage future research on keyphrase prediction to widen the evaluation of their approach.

\section{Limitations}

This study focuses on the homogeneity of keyphrase prediction models on scientific documents only in the computer science domain. As more training and evaluation resources are made available for this task such as press articles~\citep{gallina_kptimes_2019}, biomedical documents~\citep{houbre_large-scale_2022} and legal documents~\citep{salaun_europa_2024}, additional experiments on these domains would have strengthened our results.

Another limitation of our study is that the size of our evaluation collections is not big enough to allow us to verify the statistical significance of our results. Future work on how to enlarge such collections should be conducted.

\section{Ethical Statement}
This work only uses data from publicly available resources. As the reformulations of the articles in the first evaluation set are considered transformations, even if the original article were to be copyright protected, the reformulations fall under the exceptions from the 2019/79 EU guideline on using copyright content in text and data mining for research purposes. This makes them fully distributable to the community as long as they are not used for commercial purposes. We therefore share them under the Creative Commons Non Commercial use license CC-BY-NC v4.0.

\section{Bibliographical References}\label{sec:reference}

\bibliographystyle{lrec2026-natbib}
\bibliography{custom}

\end{document}

%% file: Tables/eval_data_stats.tex
\begin{table*}[ht]
    \centering
    \begin{tabular}{clc|cccc}
    \midrule
    Dataset & Version      & \#Documents   & \%P  & \%R  & \%M  & \%U \\
    \midrule
                   & Original    & 500        & 78.7 & 9.9  & 6.5  & 4.9 \\
    Inspec         & Back-translated & 170        & 69.8 & 15.5 & 8.4  & 6.4 \\
                   & Paraphrased   & 110        & 65.6 & 17.9 & 8.6  & 7.9 \\
       
    \midrule
    KP20k          & Original     & 20 000     & 58.4 & 10.8 & 17.2 & 13.6 \\
                   & CP 50\%     & 225        & 49.2 & 6.8  & 20.6 & 23.3 \\

    \midrule
    \end{tabular}
    \caption{Datasets statistics. The "CP 50\%" version of kp20k's test set lists the documents from the pairs with a Hooper's score on their author assigned keyphrases of at least 50\%. }
    \label{ch6_tab:eval_data_stats}
\end{table*}

%% file: Tables/inspec_results.tex
\begin{table*}[ht!]
    \centering
    \begin{tabular}{clcccc}
    \toprule

   & Model & CP Hooper & CP Rodgers &  \% Gen.  & Cor. ROUGE \\
    
    \midrule%
   & CopyRNN         & 46,2          & 63,4          & 28,2          & 0,12$^{\star}$ \\
   & CorrRNN         & 49,0          & 64,9          & 28,8          & 0,18$^{\star}$ \\
   & TG-Net          & 45,7          & 62,6          & 28,0          & 0,09 \\
 \parbox[t]{2mm}{\multirow{3}{*}{\rotatebox[origin=c]{90}{gen.}}}   & TGRF            & \textbf{57,6} & \textbf{69,0} & 11,7          & 0,26$^{\star}$ \\
   & BART            & 37,9          & 56,3          & 26,6          & 0,20$^{\star}$ \\
   & BART-60p        & 37,3          & 55,1          & 26,1          & 0,22$^{\star}$ \\
   & One2Set         & 29,6          & 48,8          & 40,2          & 0,24$^{\star}$ \\
   & One2Set-60p     & 48,2          & 61,4          & \textbf{41,1} & 0,37$^{\star}$ \\
    \midrule
 \parbox[t]{2mm}{\multirow{2}{*}{\rotatebox[origin=c]{90}{ext.}}}   & MultipartiteRank& 46,4          & 61,2          &    0          & 0,40$^{\star}$ \\
   & TF-IDF          & 49,3          & 62,2          &    0          & \textbf{0,48$^{\star}$} \\

    \bottomrule
    
    \end{tabular}
    
    \caption{ Homogeneity scores on pairs using reformulations on the Inspec test set. Top part concerns keyphrase generation models while the bottom part concerns keyphrase extraction methods.
    }
    \label{tab:inspec_results}
\end{table*}

%% file: Tables/kp20k_results.tex
\begin{table}[h]
    \centering
    \begin{tabular}{clccccc}
    \toprule

   & Model & CP Hooper & CP Rodgers  \\
    
    \midrule

   & CopyRNN          & 12,5             & 20,3 \\         
   & CorrRNN          & 12,0             & 18,6 \\        
   & TG-Net           & 13,8             & 20,7 \\         
 \parbox[t]{2mm}{\multirow{3}{*}{\rotatebox[origin=c]{90}{gen.}}}   & TGRF             & 11,7             & 16,6 \\
   & BART             & 10,4             & 19,4 \\         
   & BART-60p         & 14,0             & 21,4 \\        
   & One2Set          & 14,1             & 22,4 \\         
   & One2Set-60p      & \textbf{16,5}    & \textbf{22,5} \\    
    \midrule   
 \parbox[t]{2mm}{\multirow{2}{*}{\rotatebox[origin=c]{90}{ext.}}}   & MultipartiteRank & 7,2              & 13,5 \\         
   & TF-IDF           & 6,3              & 10,3 \\         

    \bottomrule
    \end{tabular}
    \caption{Results on the pairs from KP20k with a Hooper's CP $\geq 50~\%$.}
    \label{tab:kp20k_results}
\end{table}